\pdfoutput=1

\documentclass[journal,comsoc]{IEEEtran}
\usepackage[T1]{fontenc}
\usepackage{epstopdf}
\usepackage{times}
\usepackage{epsfig}
\usepackage{graphicx}
\usepackage{amsmath}
\usepackage{amssymb}
\usepackage{subfigure}
\usepackage{multirow}
\usepackage{enumerate}
\usepackage{booktabs}
\usepackage{color,xcolor}
\usepackage{enumerate}
\usepackage{multirow}
\usepackage[keeplastbox]{flushend}

\usepackage{graphicx} 
\usepackage{epstopdf}

\usepackage{verbatim}

\ifCLASSINFOpdf
\else
\fi
\usepackage{amsmath}
\usepackage[cmintegrals]{newtxmath}
\begin{document}
%
\title{Utility-Oriented Underwater Image Quality Assessment Based on Transfer Learning}
%
%
%

\author{Weiling~Chen,~\IEEEmembership{Member,~IEEE,}
		Rongfu~Lin,
		Honggang~Liao
        Tiesong~Zhao,~\IEEEmembership{Senior Member,~IEEE,}
        Ke~Gu,~\IEEEmembership{Member,~IEEE,}
        and~Patrick Le~Callet,~\IEEEmembership{Fellow,~IEEE}

\thanks{W. Chen, R. Lin and H. Liao are with Fujian Key Lab for Intelligent Processing and Wireless Transmission of Media Information, Fuzhou University, Fuzhou 350108, China (E-mails: \{weiling.chen, n191127021, 211127119\}@fzu.edu.cn).}
\thanks{T. Zhao is with Fujian Key Lab for Intelligent Processing and Wireless Transmission of Media Information, Fuzhou University, Fuzhou 350108, China and Peng Cheng Laboratory, Shenzhen 518000, China. (E-mail: t.zhao@fzu.edu.cn).}
\thanks{K. Gu is with the Faculty of Information Technology, Beijing University of Technology, Engineering Research Center of Intelligent Perception and Autonomous Control, Ministry of Education, Beijing Key Laboratory of Computational Intelligence and Intelligent System, Beijing Artificial Intelligence Institute, Beijing 100124, China (E-mail: guke.doctor@gmail.com).}
\thanks{Patrick Le Callet is with \'{E}quipe Image, Perception et Interaction, Laboratoire des Sciences du Num\'{e}rique de Nantes, Universit\'{e} de Nantes, France (E-mail: patrick.lecallet@univ-nantes.fr).}}

\maketitle

\begin{abstract}
The widespread image applications have greatly promoted the vision-based tasks, in which the Image Quality Assessment (IQA) technique has become an increasingly significant issue. For user enjoyment in multimedia systems, the IQA exploits image fidelity and aesthetics to characterize user experience; while for other tasks such as popular object recognition, there exists a low correlation between utilities and perceptions. In such cases, the fidelity-based and aesthetics-based IQA methods cannot be directly applied. To address this issue, this paper proposes a utility-oriented IQA in object recognition. In particular, we initialize our research in the scenario of underwater fish detection, which is a critical task that has not yet been perfectly addressed. Based on this task, we build an Underwater Image Utility Database (UIUD) and a learning-based Underwater Image Utility  Measure (UIUM). Inspired by the top-down design of fidelity-based IQA, we exploit the deep models of object recognition and transfer their features to our UIUM. Experiments validate that the proposed transfer-learning-based UIUM achieves promising performance in the recognition task. We envision our research provides insights to bridge the researches of IQA and computer vision.
\end{abstract}


%
\IEEEpeerreviewmaketitle

\section{Introduction}
\IEEEPARstart{I}{mages} play an important role in daily life and work. During its acquisition, transmission, storage, and display, the noises are inevitable and could degrade the quality of images. Traditional Image Quality Assessment (IQA) is developed to automatically assess the perceptual quality of images including fidelity-oriented IQA and aesthetics-oriented IQA. However, with the massive growth in complex tasks, perceptual quality may not necessarily be used for the subsequent processing beyond user enjoyment. In such a case, the traditional IQA cannot be applied and thus new techniques should be developed.

\begin{figure}[t]
	\centering
	\includegraphics[height=7.3cm,width=9.0cm]{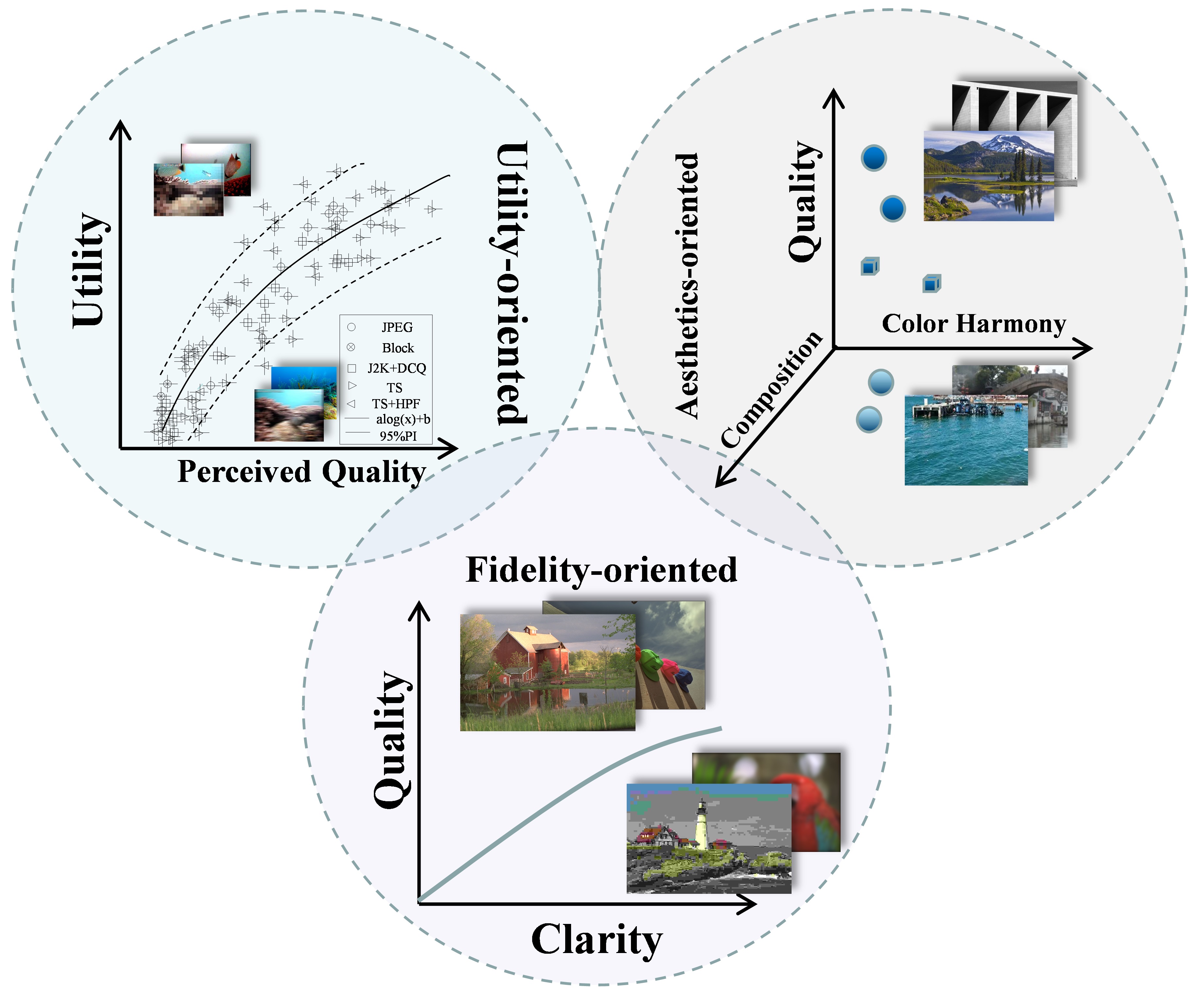}
	\caption{\small The difference among utility-oriented IQA, aesthetic-oriented IQA and fidelity-oriented IQA.}
	\label{fig:1}
\end{figure}

As shown in Fig. \ref{fig:1}, the fidelity-oriented IQA focuses on the clarity of details and textures that can be affected by the degree and type of distortion. It is the most widely used and studied IQA at present to reflect the viewability of the image. In a fidelity-oriented IQA system, the image quality is undoubtedly positively correlated with clarity, and converged as the clarity improves. In aesthetic-oriented IQA, the image gives a sense of harmony and beauty for user enjoyment. In such scenarios, image quality is positively related to composition, color harmony, \textit{etc}, in addition to fidelity. To improve the visual comforts and aesthetic perceptions of users, the aesthetic analysis of images has also been widely studied. In Fig. \ref{fig:1}, the shades of colors and different shapes are used to indicate the influence of other factors on aesthetic quality. 

Besides user enjoyment, many images have also been applied in practical scenarios like user analysis, understanding, processing. For example, underwater images are used for the exploration of marine resources. In practical scenarios, utility is the deciding factor of quality while fidelity and aesthetic are just contributing factors. Furthermore, the fidelity- and aesthetic-based quality can be collectively called perceived quality.
As described in \cite{2009Image}, the link between utility quality and perceived quality is difficult to be summarized by a single model. In \cite{rouse2011estimating}, a further relationship is mentioned, that is, perceived quality improvements correspond to smaller utility upgrades. In consequence, it is suggested that the image utility saturates more easily than perceived quality. After the saturation point, the extra enhancement of perceived quality does not benefit the utility. Furthermore, the utility is also related to the requirements of the specific task, for example, highlighted target is more desired in object detection. The utility-related characteristics are usually neglected by existing fidelity- or aesthetic-oriented IQA. In view of the above characteristics, perception-based evaluation criteria are not directly feasible in utility-oriented IQA. To evaluate the image utility, we propose a definition of utility-oriented IQA by summarizing the task descriptions of utility assessment in \cite{2009Image} as:


\emph{The quality evaluation of an image considering its utility to complete a vision-based task.}

For more specific analysis, we initialize our research in the scenario of underwater fish detection, which is a challenging task that has not yet been perfectly solved. On the basis of the Underwater Image quality database for Fish Detection (UIFD) \cite{P554}, an Underwater Image Utility Database (UIUD) is built. Specifically, we add the types of distortions that affect image quality under certain tasks, including foreground/background distortions, and tentative analysis of non-target images. Compared with several public databases, our database considers the task background from the original image selection, image degradation processing, and subjective experiment design. Then, we make good use of deep models for object recognition and transfer these features to an Underwater Image Utility Measure (UIUM). For a more intuitive explanation, the UIUM is decomposed into a main-task and a sub-task. The sub-task is pre-trained on a large-scale database for object detection to get a fish detection model, while the main-task predicts the image utility, taking advantage of key feature information obtained from the sub-task. Our major contributions are as follows:


(1) Raised and discussed the utility-oriented IQA under the scenario of object detection. We envision our research provides insights to bridge the researches of IQA and computer vision.

(2) Developed the first-of-its-kind utility-oriented IQA database. It can be utilized as a benchmark to develop and evaluate subjective methods of utility-oriented IQA.

(3) Proposed a UIUM metric based on transfer learning, which is the first successful attempt to present a utility-oriented IQA metric. Experimental results demonstrate the effectiveness of our model.

The remaining of this paper is organized as follows. Section II introduces the related work. Section III explains details about the construction of image quality database. Section IV elaborates the proposed method. In Section V, we report the experimental results. Finally, the paper is concluded in Section VI.

\section{Related work}
\subsection{Image Quality Assessment}
IQA can be divided into three categories according to its objective as described above. Fidelity-oriented IQA evaluates whether the image clearly conveys all visual information. The related studies can be classified into Natural Scene Statistics (NSS)-based methods and learning-based methods. \cite{P36} utilized an NSS model of DCT coefficients to predict image quality scores. \cite{P37} used locally normalized luminance coefficients to quantify possible losses of naturalness in images due to distortions. A small codebook was employed in \cite{P38} for a general-purpose blind IQA based on high-order statistics aggregation. \cite{P39} integrated the features of NSS derived from multiple cues to learn a multivariate Gaussian model of image patches. \cite{P40} combined feature learning and regression into one optimization process to estimate quality. A neural network-based pooling was presented in \cite{P51} to assess the global image quality with local patch qualities. To explore the relationship between fidelity and quality,  \cite{P5,P6} proposed deep learning methods. Notwithstanding the prosperity of methods, the above-mentioned ones perform well in the quality estimation of Natural Scene Images (NSIs) but fail in that of utility-oriented IQA.

Unlike fidelity-oriented IQA, aesthetics-oriented IQA prefers visual contents with reasonable layouts and visual comforts. \cite{P7} relied on artificially designed functions to improve the image aesthetics reasoning of Convolutional Neural Network (CNN). \cite{P8} proposed a two-stream CNN which considers heterogeneous and complementary aesthetic perceptual abilities respectively. \cite{P9} proposed to quantify image aesthetics by distributing it across multiple quality levels. In \cite{P10}, authors proposed a multi-reference eye inpainting Generative Adversarial Networks (GAN) approach based on an eye aesthetic dataset. \cite{P11} presented a deep multi-modal learning for aesthetic quality assessment of unmanned aerial vehicle videos. \cite{P12} employed a gated information fusion network to weight the roles of foveal vision and peripheral vision, which are key issues in image aesthetic evaluation. \cite{P13} proposed a semi-supervised deep active learning to explore the way humans perceive semantically important areas in images. Moreover, it developed a probabilistic model to incorporate the aesthetic experience of multiple users by encoding the experience of several professional photographers.
These methods focus on various visual factors and composition to enhance users experience.

The utility-oriented IQA evaluates the task-aware utility based on the richness of useful information. High-utility images help complete subsequent tasks. At present, there is few work in utility quality assessment. 
Rouse \textit{et al.} proposed the concept of utility assessment for natural images in \cite{2009Image}, where the objective evaluation was expected to be consistent with subjective judgement of usefulness. In this paper, a subjective utility quality database for natural images, referred to as the CU-Nants database, was obtained. On basis of CU-Nants database, a series of full-reference utility quality assessment methods were proposed \cite{rouse2011estimating}, \cite{2009Image},  \cite{7532327}. These methods are all built upon the hypothesis that contour degradations are consistent with  decreased perceived utility. In \cite{CNNutility}, Edward \textit{et al.} came up with a no-reference utility measurement utilizing the Oxford Visual Geometry Group’s (VGG) deep CNN for CU-Nant database. As an extent, Edward \textit{et al.} also found that highly performing utility measurements can also predict saliency for object recognition \cite{Saliencyutility}, since the utility measured is related to the contour difference between test and reference images, and the distortion in contour impacts the ability of observer to recognize object. \cite{P14} considered the influence of blur, dramatic pose variations, and occlusion on face quality. The utility evaluation was guided by the vascular structure, rather than the perceived quality of the whole retinal image in \cite{P16}. Besides, video utility evaluation algorithms have also been discussed in the context of tasks such as compression \cite{9506310}, \cite{7574903}.

Despite of these great efforts, there are still lack of a deep analysis of utility-oriented IQA and a large-scale database for underwater image utility evaluation. Most of aforementioned contributions are designed for natural images or videos. Nevertheless, underwater images are different from natural images both in terms of the statistical properties, visual characteristics and distortion types. As for underwater images, although sonar images \cite{P18} can provide acoustic information that assists underwater detection, underwater optical imaging is still critical to show more intuitive visual information and its quality assessment is different from fidelity evaluation of sonar images \cite{P19}-\cite{P21}. Therefore, the fidelity-based sonar IQA is not suitable for the utility-oriented IQA discussed in this paper. 

In response to these problems, we design a utility-based subjective experiment to construct a quality database. Then, an image quality evaluation method based on the fish detection task is proposed.

\subsection{Transfer Learning and Object Detection}
Due to the limited training samples in the existing database, IQA based on deep learning faces the problem of overfitting. Transfer learning \cite{P47} offers an effective solution by utilizing data from a related source task to improve the performance. However, there are few works on quality evaluation based on transfer learning. A transfer learning framework was described in \cite{P48}, which learned an end-to-end image quality estimator in classification or regression. In \cite{P49}, the features extracted from distorted images were transferred to the same feature space, in order to solve the problem of insufficient video contents. \cite{P50} developed an IQA architecture of multi-domain transitive transfer learning, which is associated with the ImageNet source domain, the IQA target domain, and their corresponding tasks. 
\begin{figure}[t]
	\centering
	\includegraphics[height=4.5cm,width=8.4cm]{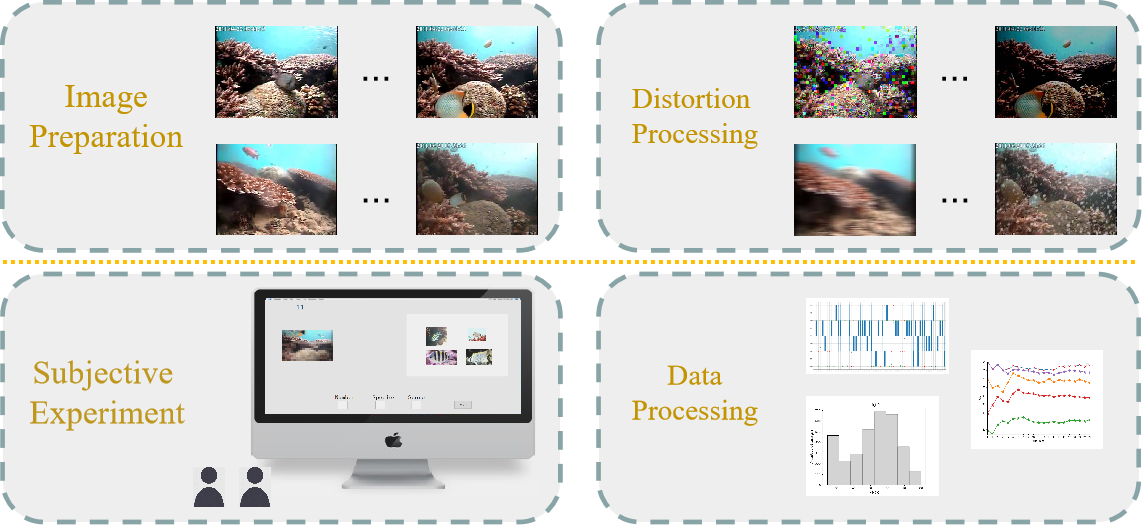}
	\caption{\small Construction of  the UIUD database.}
	\label{fig:2}
	\vspace{-0.4cm}
\end{figure}


Object detection is to find out the objects that people are interested in from images or videos, and detect their position and size. Different from the image classification task, object detection must not only solve the classification problem, but also the positioning problem. Until 2012 \cite{alexnet}, the rise of CNNs pushed the field of object detection of natural scene images to a new level.
With the development of ocean exploration and exploitation, detecting fish from underwater videos and images is of great significance for fishery resource assessment and ecological environment monitoring. In \cite{FD1}, authors proposed a novel composite fish detection framework based on a composite backbone and an enhanced path aggregation network called Composited FishNet. \cite{FD2} presented a novel dataset with 400 images of fish in the wild. By using these dataset, the state-of-the-art detection models are trained with fine-tuning strategies. \cite{FD3} proposed a deep but lightweight neural network to detect fish. 
However, due to the appalling underwater conditions, images or videos captured underwater are often with poor utility. Most fish targets are small and easily confused. It is difficult to achieve fully automatic machine recognition according to the current technology.

Inspired by these, we propose a UIUM metric based on transfer learning, which uses a trained object detection model to share the prior knowledge of key feature information for detection.

\section{Database Construction}

The ocean is unknown and changeable for human beings. A variety of fishes are important parts of the marine biosphere. Hence, monitoring the amount and species of fish is essential for regional ecological balance.
However, it is still a difficult task due to the unpredictability of species and the limited learning ability of machines. Artificial recognition is still the most reliable way under this scenario. In this paper, we establish a subjective quality database as a benchmark to develop and evaluate objective IQA methods.

A large number of general-purpose IQA databases have been built based on the standard ITU-R BT.500 \cite{P22}, such as LIVE \cite{LIVE}, TID2013 \cite{TID2013}, KonIQ-10k \cite{KONIQ} . These databases focus on visual factors such as image details, texture, color, \emph{etc}. While for aesthetic image quality databases, there are AVA \cite{PAVA}, Waterloo IAA \cite{PWA}, etc, which favor that the various visual factors in the image such as color, light, and background have a reasonable layout. Rouse \textit{et al.} developed a subjective utility database, consisting of reference and distorted versions of natural images along with corresponding subjective utility and quality ratings \cite{2009Image}. The utility scores of this database were obtained by pair-wise useful information comparison.
However, there is currently no utility-oriented underwater image quality database. In the context of underwater fish detection, we develop a first-of-its-kind database for utility-oriented IQA, with its development process summarized in Fig. \ref{fig:2}. Firstly, all source images are prepared from underwater scenes. Then, diversified types of distortions are introduced to cover typical impairments of underwater images. After that, a comprehensive subject test is conducted based on the single-stimulus test of ITU-R BT.500 \cite{P22}. Finally, all user scores are analyzed and summarized to obtain the MOS values.


\subsection{Material Preparation}

\begin{figure}[t]
	\centering
	
	\subfigure [Reference]{
		\includegraphics[height=1.7cm,width=2cm]{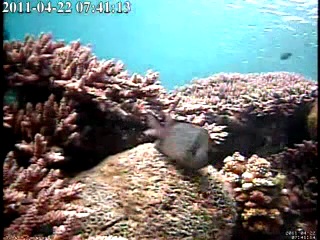}}
	\subfigure [Type1]{
		\includegraphics[height=1.7cm,width=2cm]{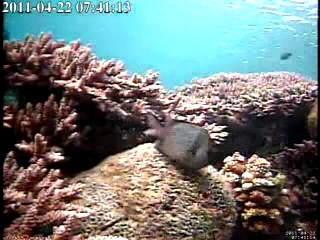}}
	\subfigure [Type2]{
		\includegraphics[height=1.7cm,width=2cm]{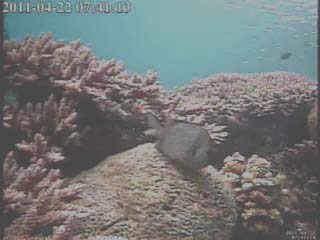}}
	\subfigure [Type3]{
		\includegraphics[height=1.7cm,width=2cm]{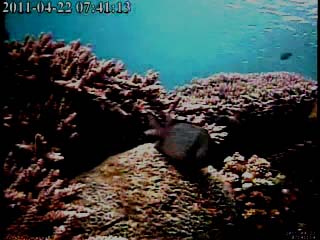}}
	\subfigure [Type4]{
		\includegraphics[height=1.7cm,width=2cm]{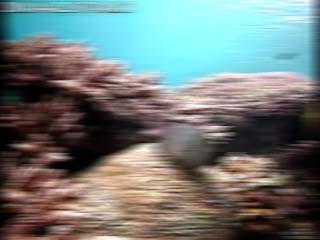}}
	\subfigure [Type5]{
		\includegraphics[height=1.7cm,width=2cm]{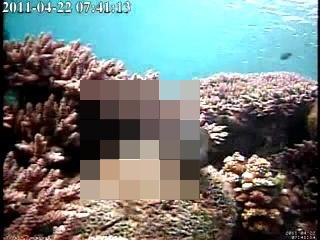}}
	\subfigure [Type6]{
		\includegraphics[height=1.7cm,width=2cm]{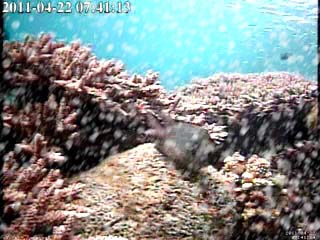}}
	\subfigure [Non-target]{
		\includegraphics[height=1.7cm,width=2cm]{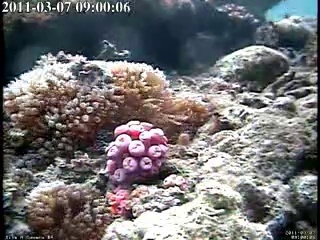}}
	
	\caption{\small Examples of the images in the UIUD database.}
	\label{fig:exm_UIQD}
	\vspace{-0.4cm}
\end{figure}

We have built a fish image quality database (\textit{i.e.}, the UIFD database) which contains 2675 fish images. This database employed the Fish4Knowledge video repository, which was taken to monitor coral reefs, to simulate the underwater environment and combine it with an underwater fish detection task. In this work, 145 images with clear fish characteristics were selected from the Fish4Knowledge video repository for the UIFD database, and each image includes 1 to 6 fishes. To simulate the complex underwater environment, five typical underwater distortions are devised, including channel distortion, contrast distortion, illumination distortion, motion blur, and ocean-snow distortion. Besides, the foreground/background distortion is also introduced to cover the impact of regional distortions on object detection. There are 4 to 5 distortion levels for most of the distortion types. Besides, we add 90 non-target images to conduct a tentative analysis. A high-fidelity image without targets is considered as a low-utility image. By introducing the foreground/background distortions as well as non-target images, the proposed database covers different utilities of underwater images.

In the proposed UIUD database, a total of 3,340 images are generated including original images and the corresponding distorted images. We present some examples of the UIUD database in Fig. \ref{fig:exm_UIQD}. For convenience, the distortion type is marked as ``Type1'' to ``Type6''. The detailed information corresponding to each type is given in Table \ref{tab:Composition}.

\begin{table}[t]
	\caption{Composition of the UIUD database.}
	\label{tab:Composition}
	\centering
	\setlength{\tabcolsep}{3mm}
	\begin{tabular}{ccc}
		\toprule
		Distortion Types	& Number	&Numbering	 \\ \midrule
		Channel Distortion	&495	&Type1		\\
		Contrast Distortion	&580	&Type2		\\
		Illumination Distortion	&580	&Type3		\\
		Motion Blur	&580	&Type4		\\
		\begin{tabular}[c]{@{}c@{}}Foreground/Background\\Distortion\end{tabular} 	&580	&Type5		\\
		Ocean-Snow Distortion	&290	&Type6					\\\midrule
		Reference Image	&\multicolumn{2}{c}{145} \\ 
		Non-Target	&\multicolumn{2}{c}{90} \\
		Total Image	&\multicolumn{2}{c}{3340} \\
		\bottomrule
	\end{tabular}
\end{table}

\subsection{Subjective Quality Evaluation}
\label{SQE}
%
%

We invited 21 subjects to conduct subjective experiments on the supplementary images except for no-target images. The number of subjects can make the experimental results reach the saturation point. All subjects have been pretrained with sufficient knowledge of underwater fishes. The collected information includes the number of fishes, the number of fish species, the time for the subject to make a judgment, and the quality of this image. The first three kinds of information are analyzed in further research. In the UIUD database, only the subjective scores are used. 

Particularly, the scoring of image quality is utility-based in this work. Traditional Absolute Category Rating (ACR) defined in ITU-R BT.500 utilizes five adjectives as categorical scales to assess image quality, which is called adjective categorical judgment methods as the right column of Table \ref{tab:differentrate} shows. The UIUD database focuses on the utility of image, \textit{i.e.}, whether or not the fish can be detected. So we develop the rating scales as given in the left column of Table \ref{tab:differentrate}. In Table \ref{tab:differentrate}, the first adjective reflects the completeness of fish information, while the second adjective establishes the identification threshold of target fish. Apart from that, the experimental environment and monitor are calibrated according to the recommendations of ITU-R BT.500 \cite{P22}.

Finally, to improve the reliability of results, we randomly sampled 5 images to form a verification set. If a subject presents significantly different scores between the verification set and those corresponding images in the subjective test, his or her score will be discarded. In detail, if the difference between two scores of the same image is more than 2, the image is regarded as a fluctuation image. In this subjective test, the average number of fluctuation images of each subject is less than 1. Thus no score has been discarded, which means that all subjects are reliable.
%

\begin{table}
	\caption{The rating criteria of UIUD and traditional subjective experiments.}
	\label{tab:differentrate}
	\begin{center}
		\begin{tabular}{c|c|c}
			\toprule
			rating &\noindent{\bf UIUD } & \noindent{\bf Traditional Databases}  \\
			\midrule
			5 & Complete and obvious &Excellent\\
			\midrule
			4 &Incomplete but uninfluential &Good\\	
			\midrule
			3 & Incomplete but identifiable &Fair\\		
			\midrule
			2 & Incomplete but distinguishable &Poor \\
			\midrule
			1 & All lost or undistinguishable &Bad\\

			\bottomrule
		\end{tabular}
	\end{center}
\end{table}

\begin{figure}[t]

	\includegraphics[width=9cm]{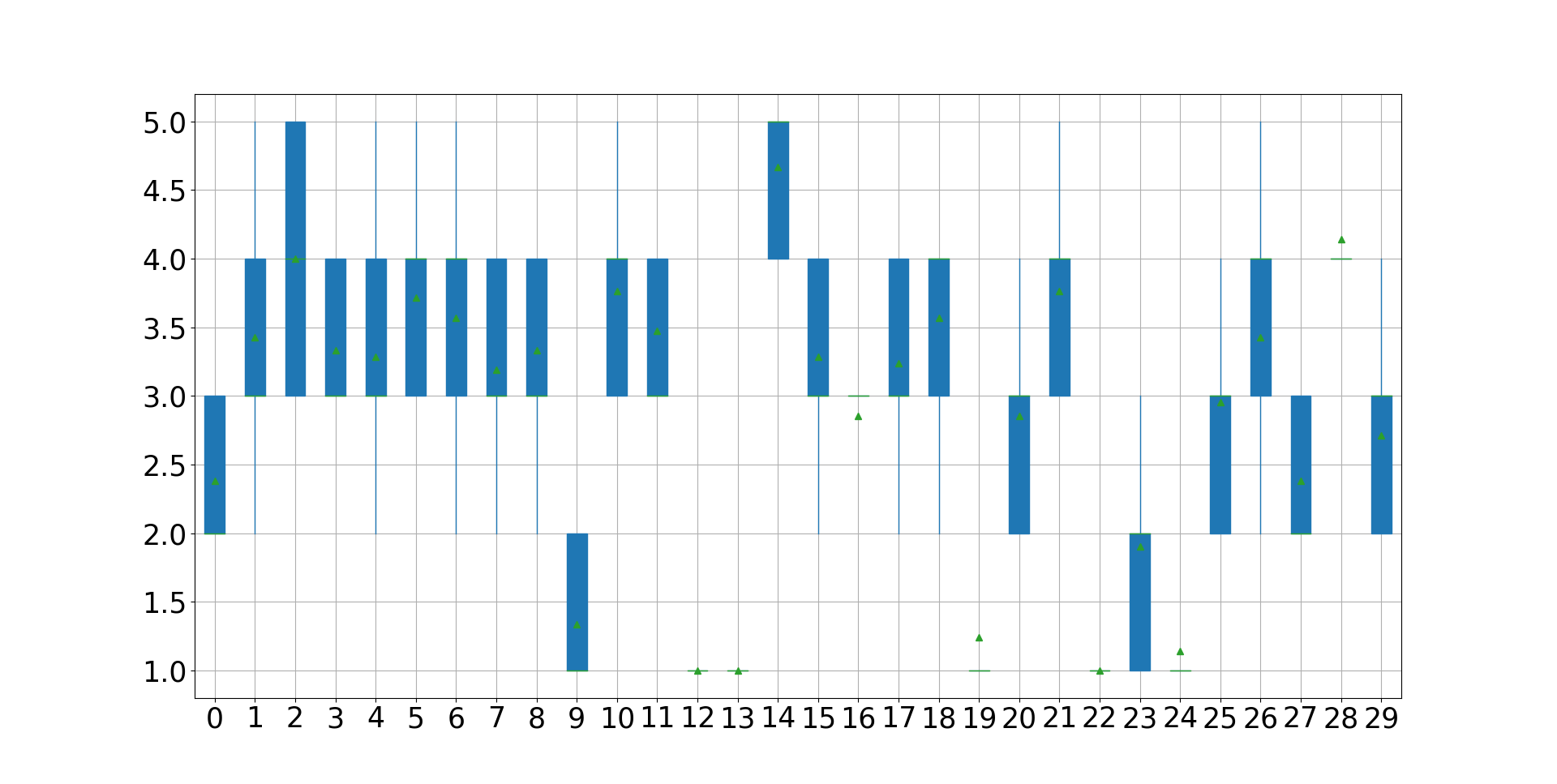}
	\caption{\small The boxplot of subjective scores. The horizontal axes corresponds to the image number, and the vertical axes corresponds to the subjective scores. The two ends of the blue rectangular box represent  the corresponding 25th and 75th percentiles of the subjective scores. The upper and lower blue horizontal lines represent the maximum and minimum values respectively.}
	\label{fig:plotbox}
\end{figure}

\begin{figure}[t]
	\centering
	
	\subfigure []{
		\includegraphics[height=3.1cm,width=4.1cm]{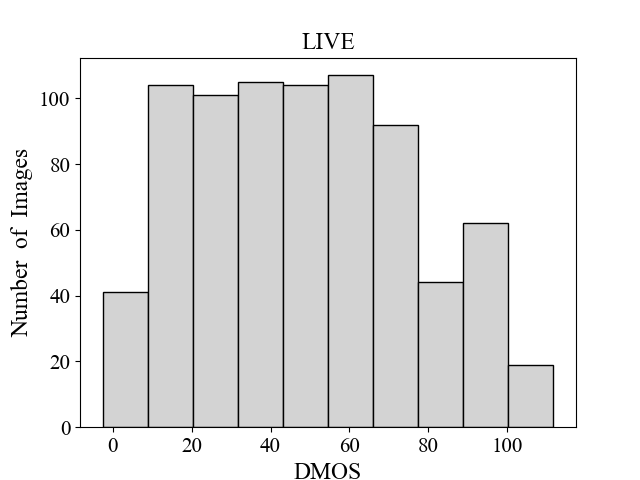}}\hspace{0mm}
	\subfigure []{
		\includegraphics[height=3.1cm,width=4.1cm]{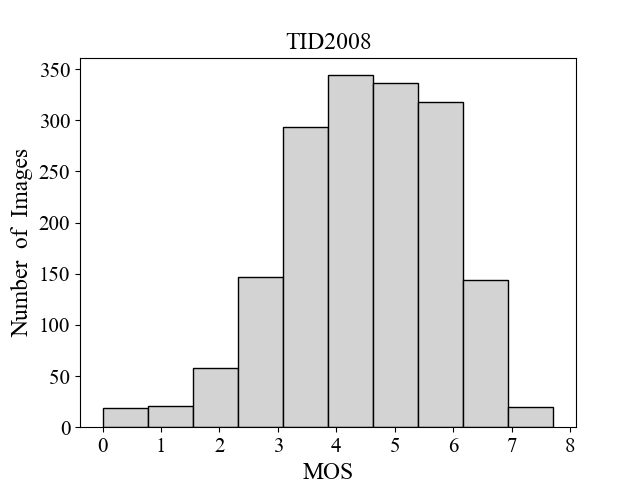}}
	\subfigure []{
		\includegraphics[height=3.1cm,width=4.1cm]{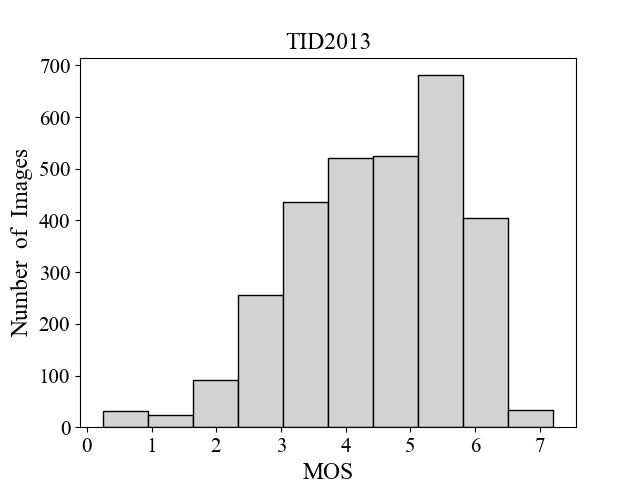}}\hspace{0mm}
	\subfigure []{
		\includegraphics[height=3.1cm,width=4.1cm]{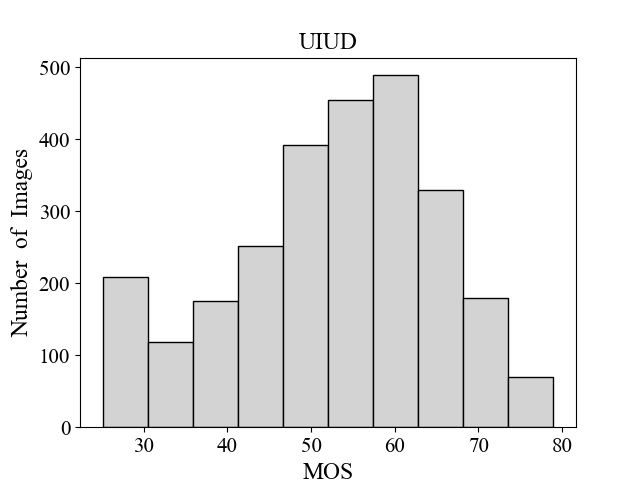}}
	\caption{\small The MOS histograms of UIUD and other public databases. (a) The histogram of LIVE. (b) The histogram of TID2008. (c) The histogram of TID2013. (d) The histogram of UIUD. }
	\vspace{-0.5cm}
	\label{histdatabase}
\end{figure}

\begin{figure*}
	\begin{center}
		\includegraphics[height=5.3cm,width=17.2cm]{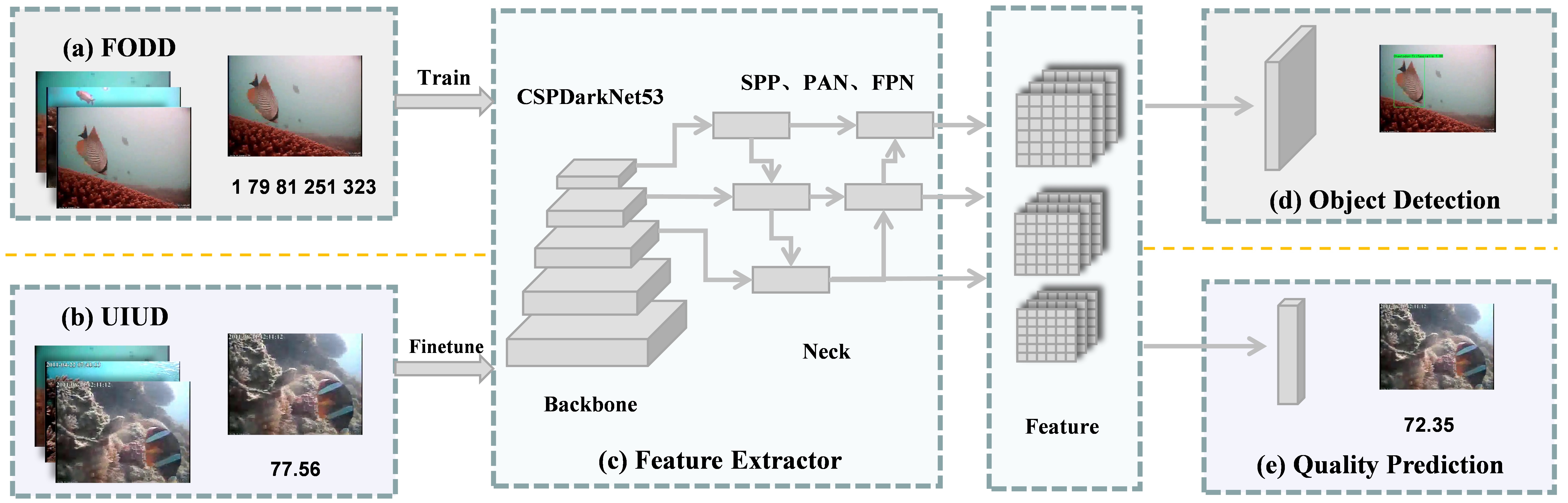}
	\end{center}
	\caption{The framework of UIUM model. The images shown in (a) are example images of FODD. The number 1 and 79, 81, 251, 323  represent the type and 2D coordinates of a fish, respectively. The images shown in (b) are example images of UIUD. The number represents the utility quality of the image. (c) is a feature extractor. (d) describes the sub-task and (e) indicates the main-task. They are two output structures used for object detection and quality regression, respectively.}
	\label{fig:framework}
\end{figure*}

\subsection{Data Processing and Analysis}
Follow the practices in \cite{P23}, we further verify the reliability of user scores obtained in Section \ref{SQE}. Outlier Coefficient (OC) is introduced to quantify the subjective agreement of the UIUD database:
\begin{equation}
\quad OC = \frac{N_\text{outlier}}{N_\text{total}},
\end{equation}
where $N_\text{total}$ denotes the total number of labeled images, and $N_\text{outlier}$ denotes the number of the images whose  interval between the 25th and 75th percentiles of subjective ratings is larger than 1. To visualize the results of OC,  Fig. \ref{fig:plotbox} shows the boxplot of subjective scores of 30 images. A subjective score is considered as an outlier when its blue rectangle is larger than 1. According to the analysis, our database achieves an OC of 5\% and thus is considered to be with high subjective agreement. Then, we process the subject rating values given by each viewer into vectors and then compute the Normalized Cross Correlation (NCC) and the Euclidean Distance (EUD) between every two vectors. The final values of NCC and EUD achieve 0.91 and 0.08, respectively. The results demonstrate the high correlation between two subjective rating vectors.


After data processing, Mean Opinion Score (MOS) values were calculated and used as the image labels. The MOS histograms of the UIUD database and other IQA databases are shown in Fig. \ref{histdatabase}. Among them, the proposed database has its own characteristics in data distribution. The data often presents Gaussian distribution without task constraints as Fig. \ref{histdatabase}(a)(b)(c) show. However, there are discontinuities according to different requirements in subjective experiments with task backgrounds. The scores increase with more fish information but saturate when the image clarity increases to a certain level. In contrast, there are fewer mid-quality images. This is consistent with the facts reflected in Fig. \ref{histdatabase}(d). There are few images with mid-quality between 30 and 45 and also few images with high quality above 65. The different distributions of scores lead to different characteristics of UIUM compared with conventional IQA approaches.

\section{Transfer learning for UIUM}
\subsection{Motivation of Framework Design}

Underwater images for fish detection have their particular features which are different from features extracted in fidelity- or aesthetic-oriented IQA. Traditional methods such as pixel-mapping-based measurements and textural features are not targeted in utility measurement.
Moreover, there is still a lack of training data in real-world applications. To address these problems, the transfer learning technique, which requires less training data, is utilized in this paper to transfer utility-based features from fish detection networks to an IQA model. 
In addition, the introduction of transfer learning is also motivated by the top-down design of perception-based, in which IQA learns various perceptual characteristics of the human visual system. As shown in Fig. \ref{fig:framework}, we design a dual-stream output structure that is decomposed into main-task and sub-task. The sub-task is trained on a fish detection database to get a fish detection model. The main-task predicts the utility-oriented quality of images.


\subsection{Network Architecture}

The proposed network includes a shared layer and a dual-stream output layer. The shared layer utilizes the YOLOv4 which consists of backbone and neck to transfer fish detection network to utility-oriented IQA. It will be used as a feature extractor for both the main-task and the sub-task.  The dual-stream output layer completes object detection and quality prediction, which can guide the realization of transfer learning.

In the shared layer, the backbone and neck are critical detectors to extract basic features for object detection. 
We employ CSPDarknet53 as the backbone for its high accuracy and low complexity. CSPDarknet53 is a combination of Yolov3 backbone network and Cross Stage Partial Network (CSPNet) \cite{P31}. CSPNet mainly solves the problem of complex computation from the perspective of network structure design. The neck further processes the important features extracted by the Backbone. In order to better extract the features concerned by the fish detection task, Spatial Pyramid Pooling (SPP) \cite{P32} module, Feature Pyramid Networks (FPN) \cite{P33} and Path Aggregation Network (PAN) \cite{P34} module of YOLOv4 are implemented. The SPP module employs different kernels for pooling and then concatenates feature maps of different scales, thus it can effectively increase the receiving range of backbone features and significantly separates the most important context features. The FPN layer conveys strong semantic features from top to bottom, while PAN transports strong positioning features from bottom to top. They work together to aggregate parameters from different backbone layers to different detection layers.

After the shared layer, three feature maps under different scales and receptive fields are obtained.  Fig. \ref{fig:framework} (d) is used for sub-task, whose loss function is generally composed of classification loss and bounding box regression loss. The output of Fig. \ref{fig:framework} (d) verifies the performance of object detection. Fig. \ref{fig:framework} (e)  is designed for main-task, \textit{i.e.}, predicting quality. The fully connected layer is exploited to achieve quality prediction. 
Experimental result shows that an optimal number of fully connected layer is 1. We connect feature maps into a one-dimensional vector. This vector will be inputted into this fully connected layer. The Mean Square Error (MSE) is employed as a loss function in the main-task, which is widely used in various regression tasks.

\subsection{Transfer Learning from Object Detection to IQA}
After building the network architecture, we exploit the shared layer and dual-stream output layer to transfer an object detection model to an IQA model. Our transfer learning process is performed in two steps: pre-training and fine-tune.

Throughout the training process, we utilize three databases, namely the UIUD database, Fish Object Detection Database (FODD) and Microsoft Common Objects in Context (COCO) \cite{P552}. The UIUD database is the image quality database established in this paper, and its label is a subjective utility quality score. FOOD and COCO are both object detection databases, and their labels are categories and coordinates. COCO is a large-scale object detection database that mainly obtains images from complex daily scenes. FODD is a database that we compiled to train the fish detection network. We denote these training datasets UIUD, FODD and COCO by $D_\text{t}$, $D_\text{f}$ and $D_\text{c}$, respectively.

\begin{figure}[t]
	\centering
	\subfigure[] {
		\includegraphics[height=3.1cm,width=4.0cm]{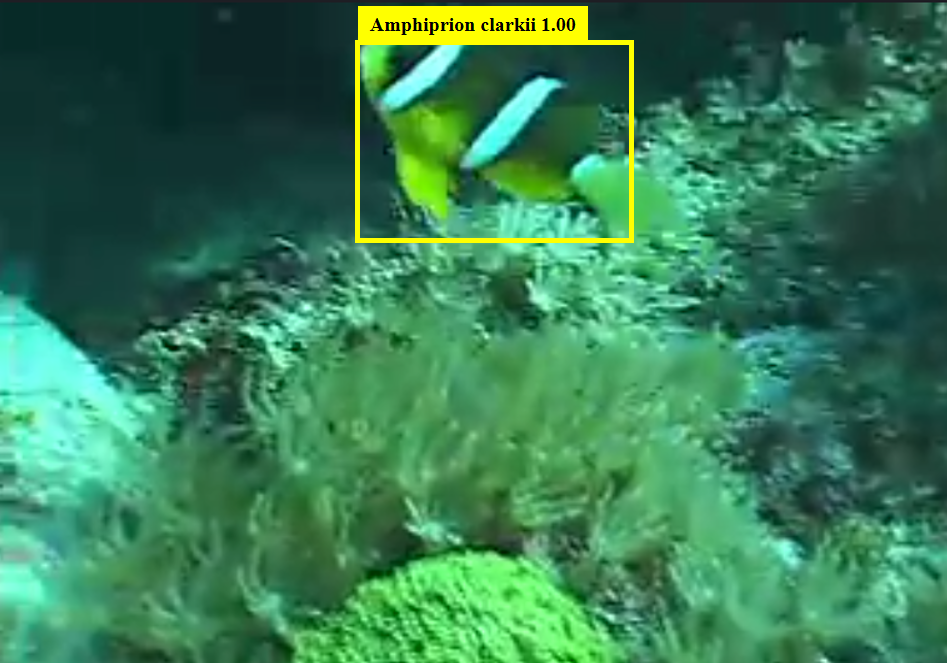}}
	\subfigure[] {
		\includegraphics[height=3.1cm,width=4.0cm]{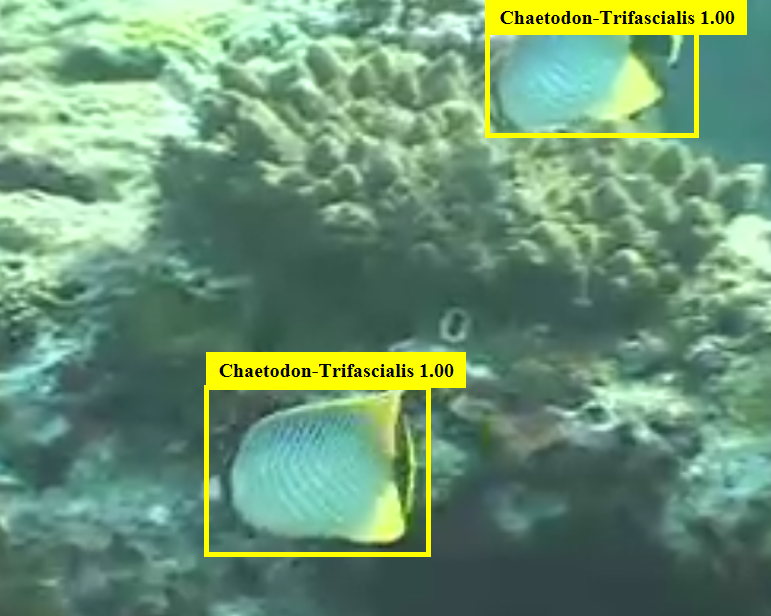}}
	\subfigure[] {
		\includegraphics[height=3.1cm,width=4.0cm]{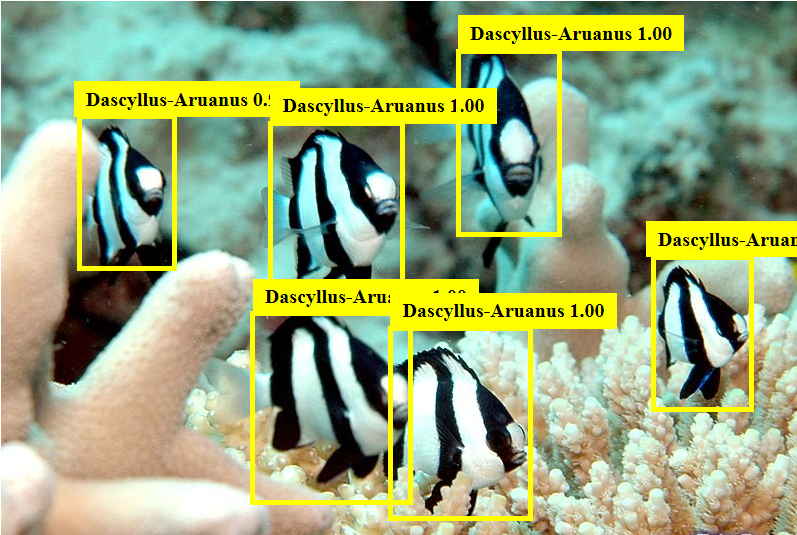}}
	\subfigure[] {
		\includegraphics[height=3.1cm,width=4.0cm]{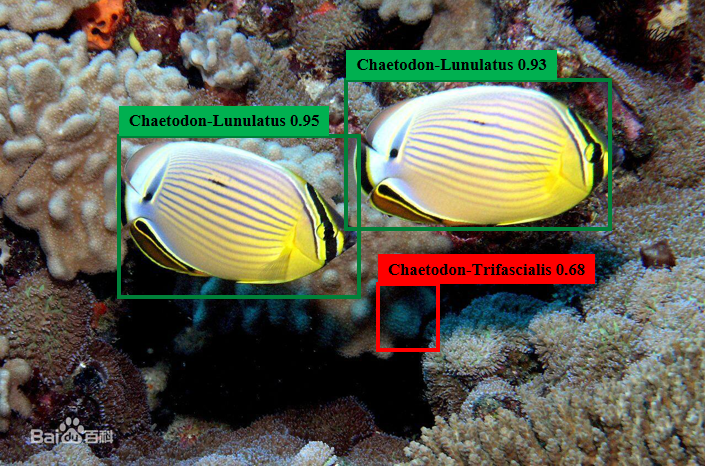}}

	\caption{\small The test results of Fish-YOLOv4. (a)(b) Images with different backgrounds. (c)(d) High-definition images.}
	\label{jiance}
	\vspace{-0.25cm}
\end{figure}
\subsubsection{Pre-training}
At this step, two training processes are carried out. We skip the network of (e) in Fig. \ref{fig:framework} and make the feature map flow into (d) in Fig. \ref{fig:framework}. The initial parameters of the shared layer are collectively denoted by $\theta$.  We first feed $\theta$ to the shared layer. The implementation process is as follows: 

\begin{equation}
\textbf{First Training: } 
\quad \theta_{c} = \mathop{\arg\min}_{\theta}(D_\text{c};L_{c}(\theta),L_{bbr}(\theta)), 
\end{equation}
where $L_{c}$ and $L_{bbr}$ are the classification loss and bounding box regression loss respectively. $\theta_{c}$ is the parameter set obtained by pre-training on COCO. Then, the network is trained on FODD to get a fish detection model. To avoid overfitting, we divide the training and testing sets by video content that each species has an appropriate ratio between the training and testing sets. The second training is implemented as:

\begin{equation}
\textbf{Second Training: } 
\quad \hat{\theta}_{c} = \mathop{\arg\min}_{\theta}(D_\text{f};L_{c}(\theta_{c}),L_{bbr}(\theta_{c})),
\label{func:3}
\end{equation}
where $\hat{\theta}_{c}$ are the optimal parameters in Fig. \ref{fig:framework} (c).

After the above trainings, we obtain a fish detection model called Fish-YOLOv4 in this paper. The mAP (mean Average Precision) of Fish-YOLOv4 can reach 0.75, which is already a relatively good performance. To further identify whether the network really learns fish detection, we check the reliability of Fish-YOLOv4 from two aspects. First, images with a different seabed background are selected as shown in Fig. \ref{jiance} (a)(b). Second, we choose high-definition images different from the first two for testing like in Fig. \ref{jiance} (c)(d). As can be seen, Fish-YOLOv4 can also successfully detect images when the style of test images is quite different. As a consequence, it can be employed as a feature extractor for the next step.

\subsubsection{Fine-tune}
At the fine-tune step, we disable the network of part (d) in Fig. \ref{fig:framework} and allow the feature map flow into part (e) in Fig. \ref{fig:framework}. The parameters of the shared layer are frozen and not optimized, and the parameters of the output layer for quality prediction are fine-tuned. Specifically, the shared layer will take the role of feature extractor to obtain utility-meaningful and utility-relevant feature maps. 
\begin{equation}
\textbf{Feature Extraction: } 
\quad F_\text{map} = \text{Fish-YOLOv4} (I;\hat{\theta}_{c}),
\end{equation}
where $I$ refers to an image in the UIUD database. $\hat{\theta}_{c}$ is the parameter obtained from pre-training. $F_\text{map}$ represents three feature maps obtained through the shared layer. In order to further verify the effectiveness of $F_\text{map}$, we perform a visualization in Fig. \ref{fmap}. The feature maps have a higher brightness in the target area, which means that higher weights are assigned to object regions. This implies that the features extracted by the shared layer are utility-related. We regard the feature maps as the inputs of the fully connected layer to get a quality score, as follows:

\begin{figure}[t]
	\centering
	\subfigure[] {
		\includegraphics[height=2.9cm,width=8.2cm]{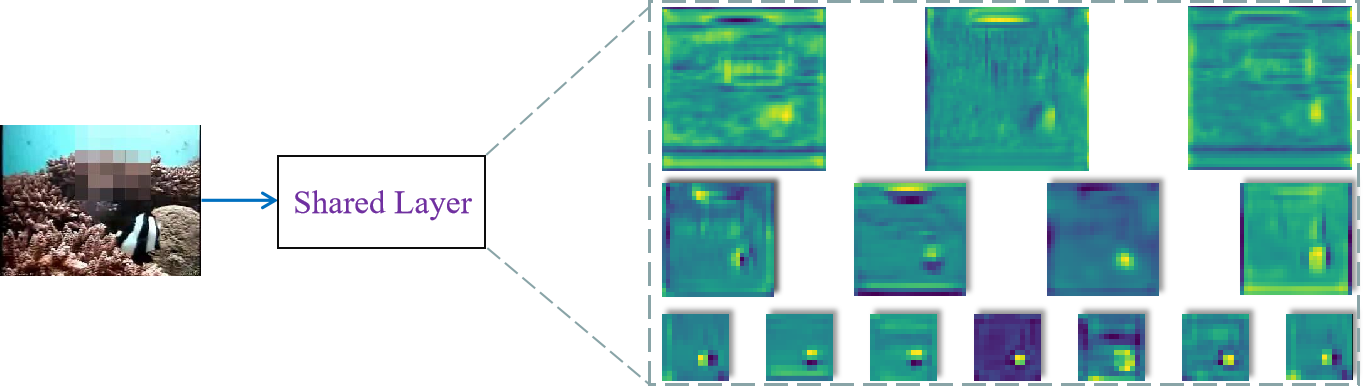}}
	\subfigure[] {
		\includegraphics[height=2.9cm,width=8.2cm]{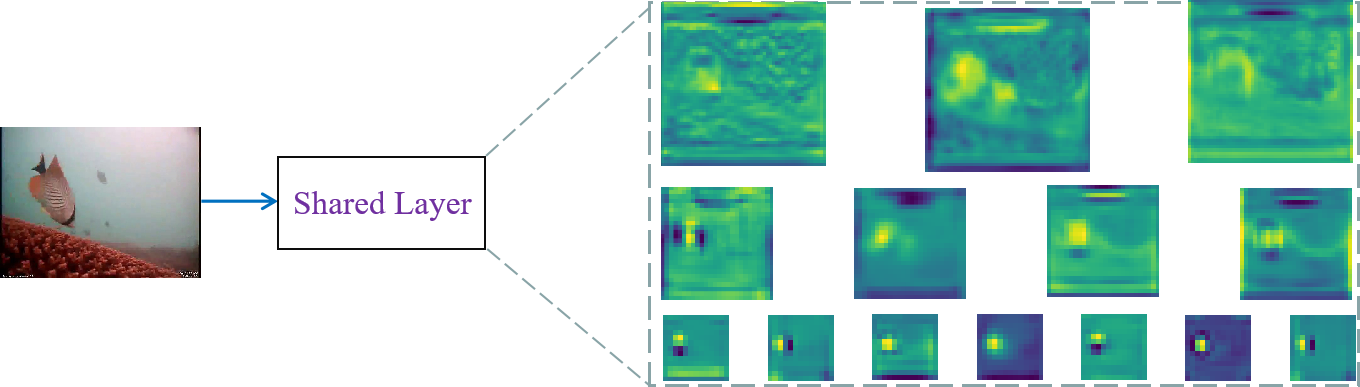}}

	\caption{\small Some examples of feature map visualization.}
	\label{fmap}
	\vspace{-0.25cm}
\end{figure}

\begin{equation}
\textbf{Quality Regression: } 
\quad {y} = \text{FC}_\text{layer}(F_\text{map};\theta_{e}),
\end{equation}
where $\text{FC}_\text{layer}$ represents the fully connected layer part (e) in Fig. \ref{fig:framework}. $y$ is the predicted quality obtained by inputting $F_\text{map}$ under the random initial parameter $\theta_{e}$. We then fine-tune the $\theta_{e}$:
\begin{equation}
\textbf{Parameter Training: } 
\quad \hat{\theta}_{e} = \mathop{\arg\min}_{\theta}  \frac{1}{N}\sum_{i=0}^{N}(\hat{y_i}-{y}_i)^{2},
\label{func:6}
\end{equation}
where $\hat{y_i}$ and ${y}_i$ represent the ground truth and predicted score of the $i$th image, respectively. After the parameter training, the optimal solution $\hat{\theta}_{e}$ of $\text{FC}_\text{layer}$ is obtained. Then UIUM is defined as:
\begin{equation}
\textbf{Quality Prediction: } 
\quad \text{Q}_{\text{utility}} = \text{UIUM}(I;\hat{\theta}_{c},\hat{\theta}_{e}),
\end{equation}
where $\text{Q}_{\text{utility}}$ is the quality of input image $I$. $\hat{\theta}_{c}$ and $\hat{\theta}_{e}$ are the optimal solutions of Eqs. (\ref{func:3}) and (\ref{func:6}), respectively.

\section{Experimental results and analysis}
In this section, we evaluate the performance of proposed UIUM model and compare it with other state-of-the-art IQA metrics.

\subsection{Experimental protocols}

\textbf{Methods for comparison.} To compare the performance comprehensively, we choose several state-of-the-art methods for comparison. They include 11 fidelity-oriented methods (BLIINDS-II \cite{P36}, BRISQUE \cite{P37}, HOSA \cite{P38}, ILNIQE \cite{P39},  WaDIQaM-NR \cite{P51}, CNN-IQA \cite{P40}, PSNR, SSIM \cite{P43}, FSIM \cite{P44}, UCIQE \cite{P45}, UIQM \cite{P46}). Among them, the WaDIQaM-NR and CNN-IQA are deep learning-based NR-IQA methods. The PSNR, SSIM and FSIM are FR IQA algorithms. In addition, the last two are underwater IQA algorithms. In particular, underwater utility-oriented IQA (NRCDM \cite{P21}) and aesthetic-oriented IQA (NIMA \cite{P466}) have also been added for comparison.

\textbf{Performance criteria.} Three commonly used criteria are chosen to calculate the correlation between the subjective and objective quality scores, which indicates the performance of IQA methods. These criteria include Spearman Rank-Order Correlation Coefficient (SRCC), Pearson Linear Correlation Coefficient (PLCC), and Kendall Rank Correlation Coefficient (KRCC).

The three metrics mentioned above can effectively evaluate the monotonicity, accuracy and consistency between the prediction quality of the algorithm and the MOS. However, these performance metrics suffer from some drawbacks. On the one hand, they do not take into account the uncertainty of subjective ratings. On the other hand, they require mapping between predicted values and subjective scores. These defects not only cause the performance metrics to be vulnerable to the quality range of the stimuli in the experiments, but also result in that the performance comparison is not performed in the real scenarios.
In order to overcome the above shortcomings, we utilize the method proposed in \cite{AUCA}, \cite{hanleymethod}, which is less dependent on the range effect and inspired by the real application without any mapping. First of all, the subjective scores need to be pre-processed as follows:
\begin{equation}
\quad 
z(i,j) = \frac{|MOS(i)-MOS(j)|}{\sqrt{\frac{var(i)}{N(i)}+\frac{var(j)}{N(j)}}},
\end{equation}
where $var(i)$ and $N(i)$ denote the variance of the subjective scores and the number of subjects of image $i$, respectively. $MOS(i)$ is the MOS value of image $i$. Then, the cumulative distribution function (cdf) of the normal distribution is employed to calculate the disparity between the qualities of image $i$ and $j$:
\begin{equation}
\quad 
p_{z(i,j)} =cdf(z)= \frac{1}{\sqrt{2\pi} }\int_{-\infty}^{z}exp( \frac{z^{2}}{2}  )dz,
\end{equation}
where the paired images are subjectively considered to be significantly different when $p_{z(i,j)}>0.95$. For significantly different pairs, the quality difference predicted by IQA model $m$ is defined as: 

\begin{equation}
\quad 
\Delta_{m}(i,j)= Q_{m}(i)-Q_{m}(j),
\end{equation}
where $Q_{m}(i)$ represents the objective score predicted by IQA model $m$ for image $i$. When $\Delta_{m}(i,j)>0.95$, the quality of image $i$ is objectively identified as significantly better than the quality of image $j$.

After the above processing, the percentage of correct recognition of qualitatively better image from the pair, which is denoted by $C_0$ (the higher the better), is employed as one performance metric.

\subsection{Performance Evaluation}

%

\begin{table}[t]
	\caption{Average Performance comparison of different IQA methods in UIUD Database.}
	\label{tab:averagepp}
	\center
	\begin{tabular}{cccccc}
		\toprule
		Methods 	& PLCC          &SRCC		    &KRCC   &Number	   &  $C_0$ \\  \midrule
		BLINDS-II  	 & 0.1100      	& 0.0520		&0.0339  & 1   & 0.4919       \\
		BRISQUE  	 & 0.1765      	& 0.1099		&0.0754  & 2   & 0.4652      	 \\
		HOSA  	 	& 0.3096       	& 0.2798		&0.1919  & 3   & 0.3662   	 \\
		ILNIQE  	& 0.2301        & 0.2398		&0.1629  & 4   & 0.4046     	 \\
		CNN-IQA  & 0.5520        		& 0.5943	&0.4181  & 5   & 0.7401    	 \\
		WaDIQaM-NR   & \underline{0.8129}        		& \underline{0.8024}		&\underline{0.6417}  & 6	 & \underline{0.8712}  	 \\
		NRCDM   &0.2562       &0.3992          &0.2598  & 7   &0.4648\\
		NIMA    &0.1088       &0.1212          &0.0801 & 8    & 0.5302  \\
		
		\midrule
		PSNR    & 0.2179	& 0.0394	&0.0420  & 9  & 0.5477 \\
		SSIM  	& {0.3740}  & 0.2912	&0.2074 & 10  & 0.6376       	\\
		FSIMc   & 0.3601    & {0.3505}		&{0.2503} & 11 & 0.6563        \\ 
		\midrule
		UCIQE    & 0.2886   &{0.1876}	&{0.1242}  & 12  & 0.4564       \\ 
		UIQM   	&{0.0124}    & 0.0155    &0.0125   & 13  & 0.4981	\\\midrule
		OURS  	& \textbf{0.8473}	&  \textbf{0.8377}   	& \textbf{0.6544} & 14 & \textbf{0.8794}\\ \bottomrule    
	\end{tabular}
\end{table}

%
%
%
%
%
%
%

\begin{figure}[t]
	\centering
	
	\includegraphics[height=3.7cm,width=8.2cm]{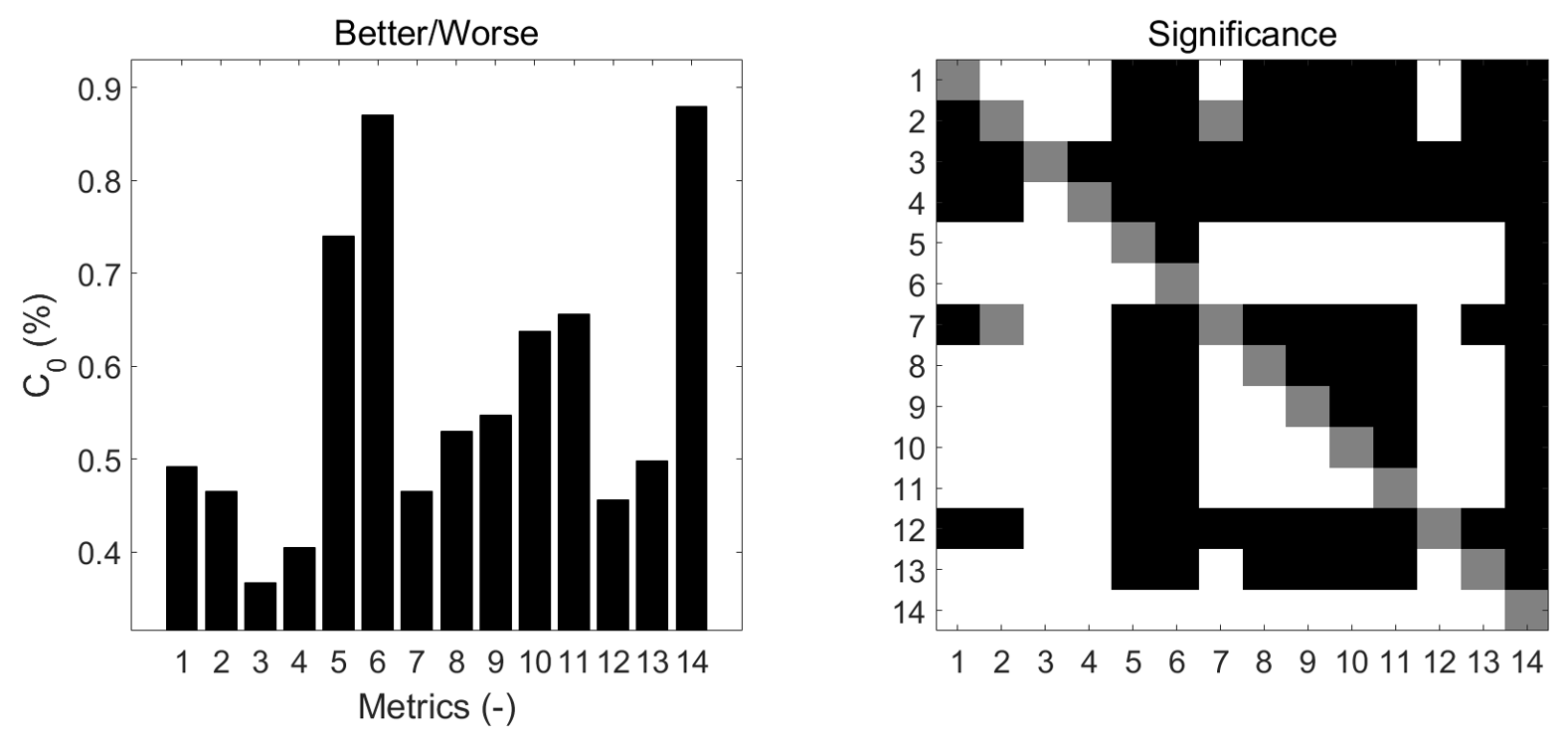}
	\caption{\small The classification ability of UIUM and other IQA methods on image quality pairs (Better/Worse) in the UIUD.}
	\label{AUCC}
	\vspace{-0.25cm}
\end{figure}

%
%
%
%

\begin{table*}[t]
	\small
	\center
	\caption{Performance on individual distortion types.}
	\label{tab:type_dif}
	\begin{tabular}{@{}cccccccccccccc@{}}
		
		\toprule
		& \multicolumn{2}{c}{Type1}  & \multicolumn{2}{c}{Type2}        & \multicolumn{2}{c}{Type3}      & \multicolumn{2}{c}{Type4} & \multicolumn{2}{c}{Type5} & \multicolumn{2}{c}{Type6} &     \\ \cmidrule(l){2-13} 
		\multirow{-2}{*}{Methods} &PLCC&SRCC&PLCC&SRCC&PLCC&SRCC & PLCC           & SRCC    & PLCC     & SRCC   & PLCC   & SRCC                  \\ \midrule
		BLINDS-II            & 0.1994           & 0.2325          & 0.2918         &0.2556         & 0.1377          &0.0087          &0.0083      &0.0478         & 0.2837          &0.2559 &0.1608  &0.1343     \\
		BRISQUE         & 0.5268           & 0.4990         & 0.4229         &0.4042          & 0.0486          & 0.0030          & 0.3873       &0.2891          &0.2671          &0.2386 &0.2551 &0.2132         \\
		HOSA               & 0.6893        & 0.7089 & {0.3701} & {0.2982} & {0.1160} & {0.0898} &  {0.1527}     &0.0621          &{0.3125}          &{0.3173}   &0.2838  &0.2470    \\
		ILNIQE         & 0.4740          & {0.6137}          & {0.5337}          & {0.5260}               & {0.2582}          & {0.2634}          & {0.3834}        & {0.3552}          & {0.1540}          & {0.0468}   &{0.2865}&0.1870              \\
		CNN-IQA            & {0.8439}     & {0.8372}               &  0.7174               & {0.7333}               & {0.6499}          & {0.6796}          & {0.6248}      & {0.6168}          & {0.4845}          & {0.5084}    &0.6339&0.6349           \\
		WaDIQaM-NR       & \underline{0.9179}                &  \textbf{0.9029}         & \underline{0.8220}          & \underline{0.8433}                &\underline{0.8270}          & \underline{0.8477}          & \textbf{0.9251}             & \textbf{0.9209}          & \underline{0.8724}          & \underline{0.8786}     &\underline{0.8561}&\underline{0.8529}      \\
	
			NSIQM            & {0.2733}     & {0.0344}               &  0.2322               & {0.1367}               & {0.1836}          & {0.1457}          & {0.2070}      & {0.2006}          & {0.0719}          & {0.0905}    &0.0667&0.0289           \\
	
	NIMA            & {0.0791}     & {0.0917}               &  0.3122               & {0.2997}               & {0.0357}          & {0.0047}          & {0.1994}      & {0.2047}          & {0.2798}          & {0.2850}    &0.0503&0.0508           \\

		\midrule
		PSNR                    & 0.7624       &0.7546     &0.3916          &0.3900                &0.3130      &0.2817      &0.3109       & {0.2262} &0.3512          &0.3199&{0.2447}&0.1653\\
		SSIM                    & 0.7901       &0.7827     &0.5293          &0.4751                &0.3540      &0.3352      &0.4420       & {0.3792} &0.3550          &0.3185&0.2069&0.1726\\
		FSIMc                    & {0.8200}       &{0.8156}     &{0.5641}          &{0.5152 }               &{0.3625}      &{0.3574}      &{0.5221}       & {0.5005} &{0.3617}          &{0.3228}&0.2132&{0.1813}
		\\\midrule
		UCIQE                    & 0.1478       &0.0469     &0.4965          &0.1933                &0.1568     &0.0059      &0.1171       & {0.0262} &0.1851          &0.1237&0.1426&0.0114
		\\UIQM                    & 0.1388       &0.1460     &0.5006          &0.4451                &0.3223      &0.3063      &0.2227       & 0.1235 &0.0828          &0.0611&0.1716&0.0498 \\ \midrule
		OURS          &\textbf{0.9213}        &  \underline{0.8950} & \textbf{0.8784} & \textbf{0.8812} & \textbf{0.8801} & \textbf{0.8745} & \underline{0.9185}  & \underline{0.9156} &  \textbf{0.8968} &  \textbf{0.8916} & \textbf{0.8669}& \textbf{0.8607} \\ \bottomrule
	\end{tabular}
\end{table*}

\begin{table}
	\caption{Time consumption (milliseconds/image) of the UIUM  method and the other mathods on the UIUD.}
	
	\label{time}
	\begin{center}
		\begin{tabular}{c|c|c|c}
			\toprule
			Method &\noindent{\bf BLINDS-II } & \noindent{\bf BRISQUE}&\noindent{\bf HOSA}\\
			
			Cost (ms) &  $6.50\times10^{3}$&   $1.39\times10^{2}$  &   $3.02\times10^{2}$\\
			
			\midrule
			
			Method &\noindent{\bf ILNIQE}& \noindent{\bf UCIQE}&\noindent{\bf UIQM}   \\
			Cost (ms) &$2.18\times10^{3}$&  $2.97\times10$&    $7.05\times10$ \\
			\midrule
			
			Method &\noindent{\bf PSNR} &\noindent{\bf SSIM}&\noindent{\bf FSIMc }    \\
			Cost (ms) &   $1.15\times10^{2}$&    $6.83$&   $9.88$\\
			\midrule
			Method&   \noindent{\bf CNN-IQA } & \noindent{\bf WaDIQaM-NR}               &\noindent{\bf OURS}\\
			Cost (ms) & $3.14\times10$&   $7.93\times10$ & $3.26\times10$\\
			\bottomrule
		\end{tabular}
	\end{center}
\end{table}

\textbf{Performance on UIUD database.} The UIUD database is divided into training and testing sets with an 80/20 split and no content overlapped. Besides, 10-fold cross-validation is utilized to evaluate the performance. Since the UIUD database is currently the first utility-oriented image quality database, we cannot perform cross-database verification. To ensure fair comparisons, the compared algorithms with deep learning are retrained and fine-tuned in the UIUD database. Table \ref{tab:averagepp} shows the average results of all methods, where the best and 2nd-best performances are highlighted with bold and underline respectively. Our method outperforms all of the state-of-the-art methods for PLCC, SRCC and KRCC evaluations. As shown in Table \ref{tab:averagepp}, only the proposed method and the WaDIQaM-NR method achieve SRCC and PLCC values above 0.8, while the performances of other algorithms are not ideal under the task of object detection. Deep network structure, feature fusion and spatial pooling methods make WaDIQaM-NR competitive. Especially, the WaDIQaM-NR network is significantly deeper than related IQA models. From this comparison, the UIUM is more relevant to utility. 
Although NRCDM evaluates images based on utility quality, its performance is less than satisfactory. This is owing to the big gap between imaging principles of acoustic image and optical image. The poor performance of NIMA results from the fact that it evaluates image content based on aesthetic appreciation. 

%
%
%

\textbf{Performance on individual distortion types.} To further understand the effect of different distortion types, we test the performance of different algorithms on each individual distortion type. The results are shown in Table \ref{tab:type_dif}, where the best and second-best results are also shown in bold font and underline, respectively. Our algorithm has high correlations in all distortion types. The other algorithms also have higher correlations in some traditional distortion types, such as contrast distortion and illumination distortion. This is obvious because these types of distortions also appear in fidelity-oriented quality evaluations.
Most algorithms perform well in transmission distortion, since the degree of distortion is highly correlated with visual perception. Furthermore, the traditional methods based on manual feature extraction fail on the remaining distortion types. The reason is that the extracted features are not suitable for underwater images, especially under task backgrounds. 

From Table \ref{tab:type_dif}, three deep learning methods have relatively superior performances due to their strong learning abilities. CNN-IQA has no advantage compared with WaDIQaM-NR and UIUM algorithms, since its network structure is relatively simple and less targeted. Quite rightly, deep learning is data-driven, we will further verify the superiority of our algorithm from the perspective of data requirements in following subsection.

\textbf{Discrimination ability for significantly different qualities.}  In order to test the discrimination ability, we calculate $C_0$ for each selected IQA methods and the UIUM. Since testing in the UIUD requires large-scale complicated computation, we randomly sampled 100 images from each distortion type and completed the test on this subset. 
For convenience, the IQA methods are numbered separately as shown in the penultimate column of Table \ref{tab:averagepp}. The last column of Table \ref{tab:averagepp} shows the exact values of $C_0$ for selected IQA methods. The results in the table indicate that UIUM (\#14) has the best performance, and the WaDIQaM (\#6) is second only to the UIUM in $C_0$. 
The same results can be obtained from the data bar  of Fig. \ref{AUCC} (left), UIUM (\#14) outperforms all other algorithms. The significance plot about $C_0$ is also shown in Fig. \ref{AUCC} (right). In this plot, black and white boxs indicate the performance of the method in the row is significantly lower and higher than the method in the column, respectively. The gray box reflects the similar performance. 
It can be concluded from the box plot and table that the UIUM is significantly better at distinguishing difference in utility than other IQA methods.

\begin{figure}[t]
	\centering
	\subfigure[M:73.56  T:72.33] {
		\includegraphics[height=2.1cm,width=2.5cm]{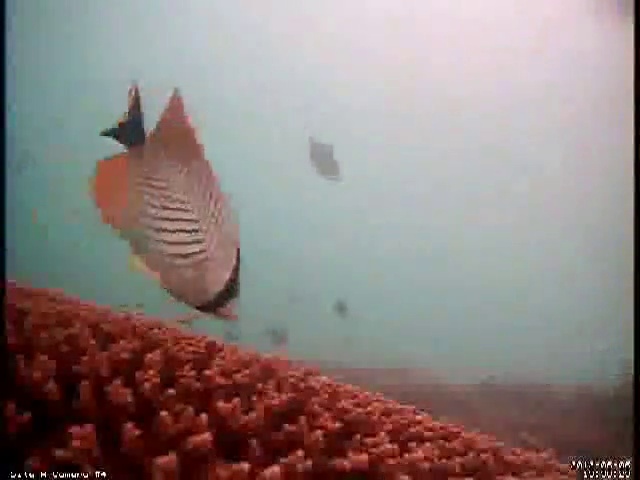}}
	\subfigure[M:70.48  T:72.57] {
		\includegraphics[height=2.1cm,width=2.5cm]{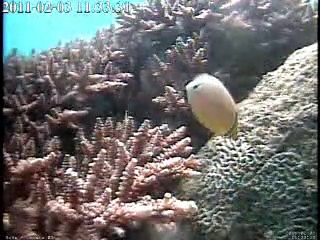}}
	\subfigure[M:77.61 T:73.54] {
		\includegraphics[height=2.1cm,width=2.5cm]{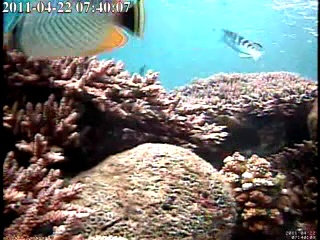}}

	\subfigure[M:69.28  T:61.19] {
		\includegraphics[height=2.1cm,width=2.5cm]{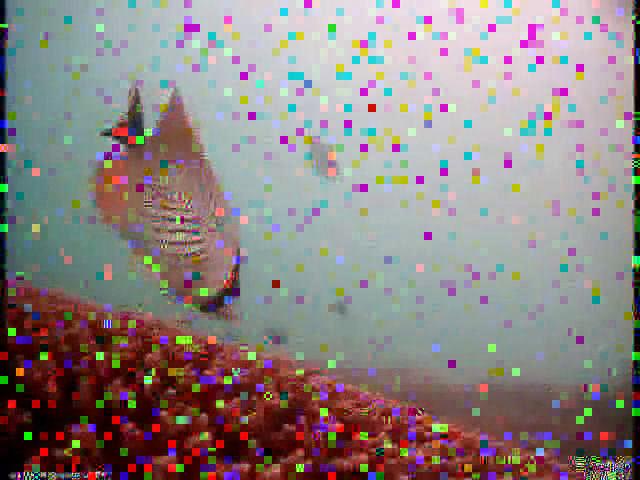}}
	\subfigure[M:74.76 T:76.84] {
		\includegraphics[height=2.1cm,width=2.5cm]{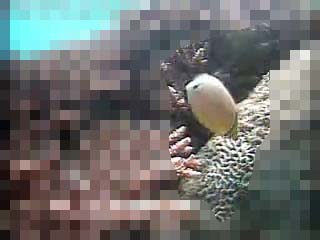}}
	\subfigure[M:75.24 T:76.03] {
		\includegraphics[height=2.1cm,width=2.5cm]{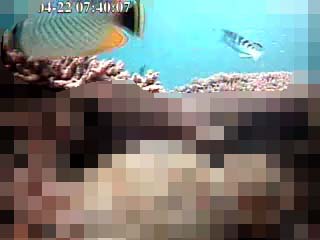}}

	\caption{\small UIUM scores for several examples that illustrate the good performance of the proposed method. M and T represent the subjective MOS value and the predicted score of UIUM, respectively.}
	\label{fig:111}
	
\end{figure}

\textbf{Intuitive Comparison.} To visually illustrate the performance of UIUM, its prediction scores and the subjective MOSs are directly compared in Fig. \ref{fig:111}. Fig. \ref{fig:111}(a) to Fig. \ref{fig:111}(c) are the reference images, while Fig. \ref{fig:111}(d) to Fig. \ref{fig:111}(f) are the corresponding distorted images. The predictions made by UIUM have a high correlation with MOS values. Another key difference between UIUM and traditional IQA is shown here is that the quality score of the original image is not necessarily higher than that of the severely degraded images, such as Fig. \ref{fig:111}(b) and Fig. \ref{fig:111}(e).

\textbf{Computation Time Comparison.} To measure the time complexity of each algorithm, UIUM and all other algorithms are run on the UIUD database for testing. These computational cost tests are conducted on the MATLAB R2019a and Pytorch software platform on a computer with a 3.98 GHz CPU, 16.00 GB of RAM and an RTX2070 graphics card. The average time consumption of each method is tabulated in Table \ref{time}. 
The first two lines are traditional NR-IQA, the third line is traditional FR-IQA, and the last line is NR-IQA based on deep learning. SSIM and FSIMc have obvious advantages in speed, but they are both methods of manually extracting features, which are difficult to achieve robust performance due to the limitation of fixed features. In the deep learning methods, the calculation time of UIUM is almost the same as that of CNN-IQA, while UIUM has great advantages over CNN-IQA in performance. WaDIQaM-NR is 4\% worse than UIUM in performance, which is the closest to UIUM among all methods. It also lags behind us in computational efficiency. In general, UIUM has great advantages in performance and calculation time.

\begin{table}[t]
	\caption{Ablation Study.}
	\label{tab:Ablation}
	\center
	\setlength{\tabcolsep}{5mm}
	\begin{tabular}{cccc}
		\toprule
		Methods 			& PLCC         		&SRCC		&KRCC    	\\  \midrule
		ODUQA  	 		& 0.7529        		& 0.7549		&0.5619        	 \\
		UIUM  	 		& 0.8473        		& 0.8377		&0.6544        	 \\
		\bottomrule  
	\end{tabular}
\end{table}

\subsection{Ablation Study}
To further evaluate the impact of transfer learning, we conduct an ablation experiment. In the UIUM, the pre-trained weights are firstly fed in COCO. Then the model is trained on FODD to obtain a fish detection model. Finally, we fix the parameters of backbone and neck, and fine-tunes the fully connected layer on UIUD database. In the ablation experiment, we define an Object Detection Utility-Oriented Quality Assessment (ODUQA). The framework of ODUQA is the same to UIUM except that the ODUQA is trained on a normal object detection model, not a fish detection model. 
These two methods are tested separately in the UIUD database. The results in Table \ref{tab:Ablation} shows that our method achieves higher performance. It demonstrates that the features extraction module is able to precisely capture the useful information for the target domain (\textit{i.e.}, utility-oriented IQA) from the source domain (\textit{i.e.}, fish detection).

\subsection{Advantages of Transfer Learning on Database}
In this work, transfer learning makes full use of the labeled data of source task, which enable better performance of a new task with less labeled data. However, data annotating is boring and expensive.
Therefore, we discuss the performance of UIUM and WaDIQaM-NR on smaller data sets in this section. The reason for choosing WaDIQaM-NR is its high performance on the complete UIUD database, and it is also an excellent deep learning-based NR-IQA method in recent years. After randomly deleting part of the reference images and their corresponding distorted images, we got two databases with 1007 and 2015 images, respectively. In this section, we name the databases according to the number of images, which are UIUD-3340, UIUD-2015, and UIUD-1007. The proportion of testing and training sets is the same as above.

\begin{figure}[t]
	\vspace{0.25cm}
	\centering
	\includegraphics[width=9cm]{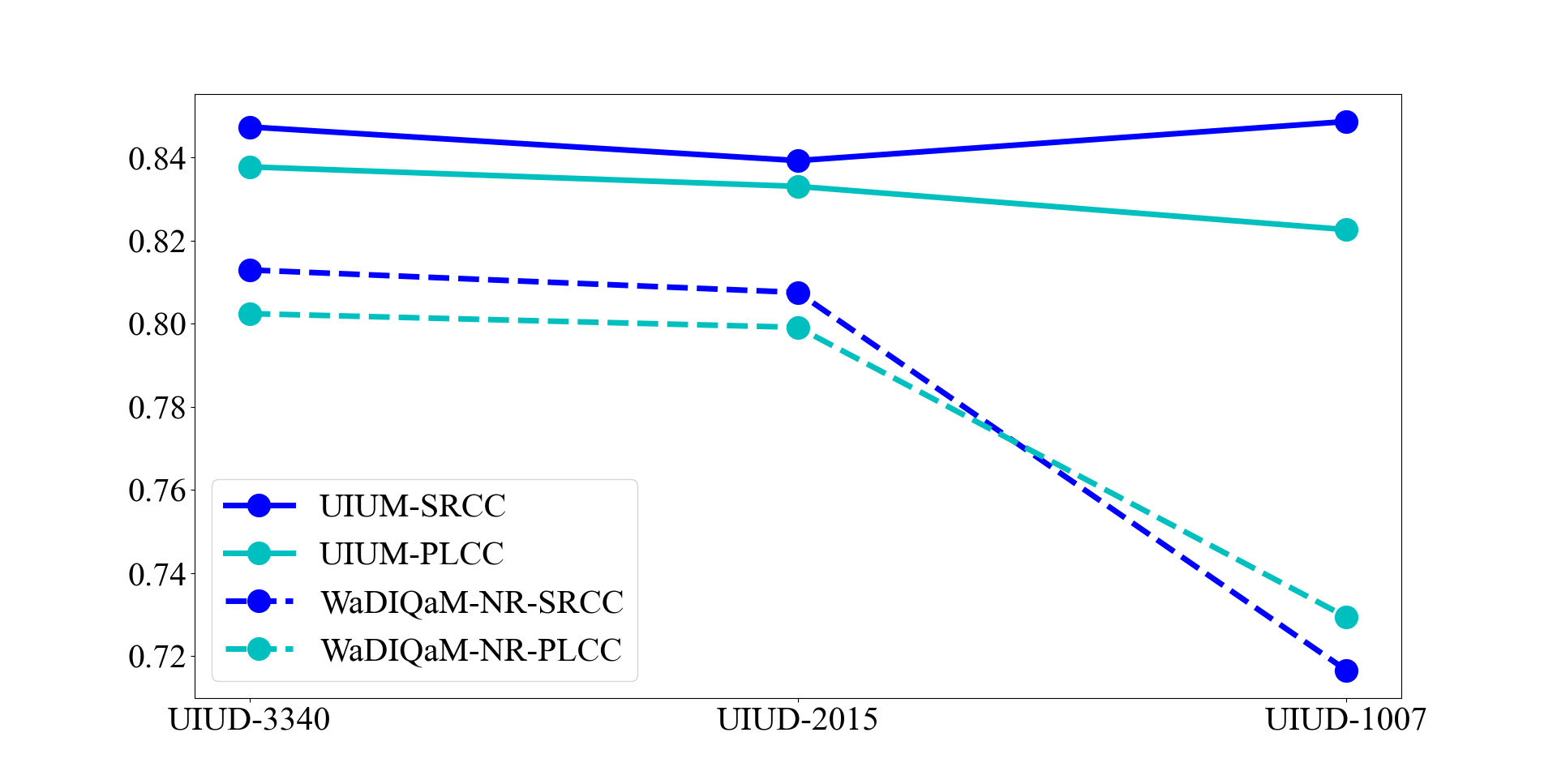}
	\caption{\small The performances of UIUM and WaDIQaM-NR on UIUD with different image numbers.}
	\label{smalldata}
\end{figure}

\begin{figure}[t]
	\vspace{0.25cm}
	\centering
	\includegraphics[width=9.0cm]{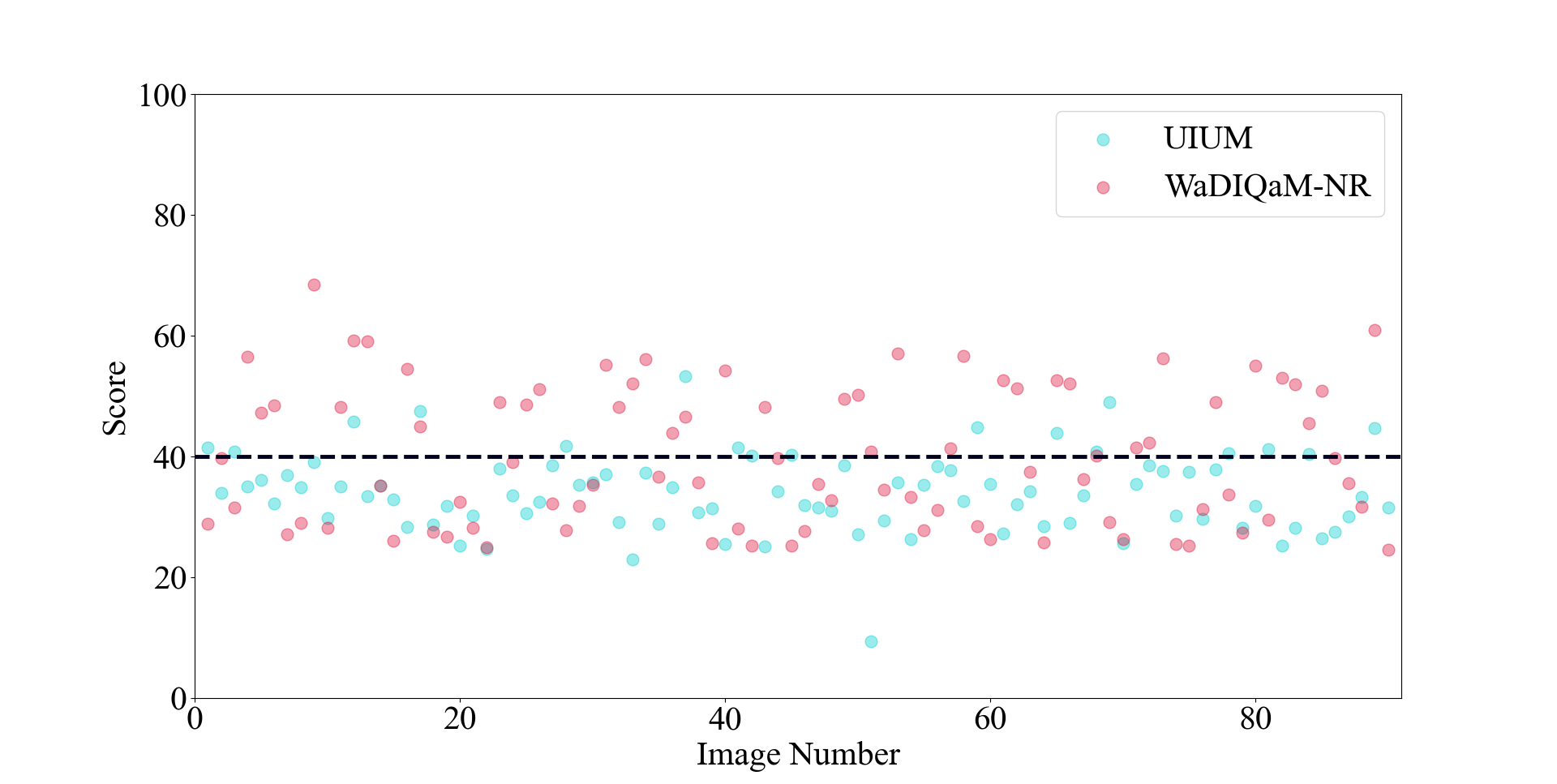}
	\caption{\small Scatter plot of prediction scores of UIUM and WaDIQaM-NR on non-target images. The horizontal axes correspond to the image number, and the vertical axes represent the image scores.}
	\label{nobojectt}
\end{figure}

The performances of UIUM and WaDIQaM-NR on the UIUD database with different image numbers are shown in Fig. \ref{smalldata}. 
When the number of images changes, the performance of UIUM remains the same.
However, the performance of WaDIQaM-NR drops significantly. When the number of images becomes 1007, its SRCC and PLCC drop to 10\% and 8\%, respectively. This fully proves that our algorithm employing transfer learning achieves better performance with less training data. Furthermore, the performance of WaDIQaM-NR starts to saturate on UIUD-2015. Specifically, the optimal SRCC and PLCC that WaDIQaM-NR can achieve in this task is about 0.8, while the UIUM yields a 5\% performance gain 
In consequence, UIUM achieves higher performance in a smaller database, and its final performance is also better than WaDIQaM-NR.

\subsection{Analysis of Non-Target Images}
We have added 90 non-target images to UIUD. The reason is that the task cannot be completed with a distortion-free and clear but targetless image. Therefore, we tentatively verify the non-target images. The result is shown in the Fig. \ref{nobojectt}. It can be found that UIUM can define most targetless images as low-quality images below 40 points, and the prediction of WaDIQaM-NR for these images is more volatile, even with predictions of more than 60 points. In the future, we will add non-target images and further analyze the content in combination with the background of specific tasks.

\section{CONCLUSIONS}

Nowadays, the IQA has been a popular vision task in quality monitoring and optimization during acquisition, transmission, enhancement, \textit{etc}. In this work, we envision another application of IQA, namely utility-oriented IQA, which associates image quality with its utility in a vision-based task. We conduct our work in the context of fish detection, since it is of great significance for underwater exploration and it is difficult to achieve automatic analysis according to the current state of the art. We firstly develop a database consists of representative images for fish detection and their typical distorted versions, named UIUD. To our knowledge, the UIUD database is the first utility-oriented image quality database. Then we propose a UIUM algorithm to achieve utility measurement. We extract utility-related features by employing transfer learning, which transfers characteristic features in fish detection to utility-oriented IQA with a shared layer. The proposed framework can be easily extended to general object detection tasks. 
We hope our research can initiate the quality evaluation in general computer vision tasks and expand the field of IQA. The UIUD database and the UIUM model will be made publicly available to facilitate reproducible research.

\section*{Acknowledgment}
UIFD was published in 2021 IEEE/CIC international conference on communication in China.  Here we reorganize its structure to evaluate the utilities of underwater images.

\appendices

\ifCLASSOPTIONcaptionsoff
  \newpage
\fi



\bibliographystyle{IEEEtran}

\bibliography{ref1}

%
%
%

%

%




\end{document}